\documentclass[letterpaper, 10 pt, conference]{ieeeconf}  

\IEEEoverridecommandlockouts                              

\usepackage{graphics} 
\usepackage{amsmath} 
\usepackage{amssymb}  
\usepackage{wrapfig}
\usepackage{graphicx}
\usepackage{booktabs}
\usepackage[dvipsnames]{xcolor}

\DeclareMathOperator*{\argmax}{arg\,max}

\usepackage[colorlinks=true]{hyperref}
\newcommand{\mypara}[1]{\par\vspace*{1.5mm}\noindent\textbf{{#1}}}

\usepackage{soul}

\title{\LARGE \bf
SSCNav: Confidence-Aware Semantic Scene Completion for \\ Visual Semantic Navigation
\vspace{-3mm}
}

\author{Yiqing Liang \quad Boyuan Chen \quad Shuran Song\\
Columbia Unviersity\\
\url{https://sscnav.cs.columbia.edu/}
 \thanks{This work was supported in part by the Amazon Research Award, Columbia School of Engineering and National Science Foundation under CMMI-2037101.}
}

\begin{document}

\maketitle
\thispagestyle{empty}
\pagestyle{empty}

\begin{abstract}

This paper focuses on visual semantic navigation, the task of producing actions for an active agent to navigate to a specified target object category in an unknown environment. 
To complete this task, the algorithm should simultaneously locate and navigate to an instance of the category. In comparison to the traditional point goal navigation, this task requires the agent to have a stronger contextual prior to indoor environments. We introduce SSCNav, an algorithm that explicitly models scene priors using a confidence-aware semantic scene completion module to complete the scene and guide the agent's navigation planning. Given a partial observation of the environment, SSCNav first infers a complete scene representation with semantic labels for the unobserved scene together with a confidence map associated with its own prediction. Then, a policy network infers the action from the scene completion result and confidence map. Our experiments demonstrate that the proposed scene completion module improves the efficiency of the downstream navigation policies. Code and data: \url{https://sscnav.cs.columbia.edu/}

\end{abstract}


\section{INTRODUCTION}

\begin{figure}
\centering
\includegraphics[width=0.97\linewidth]{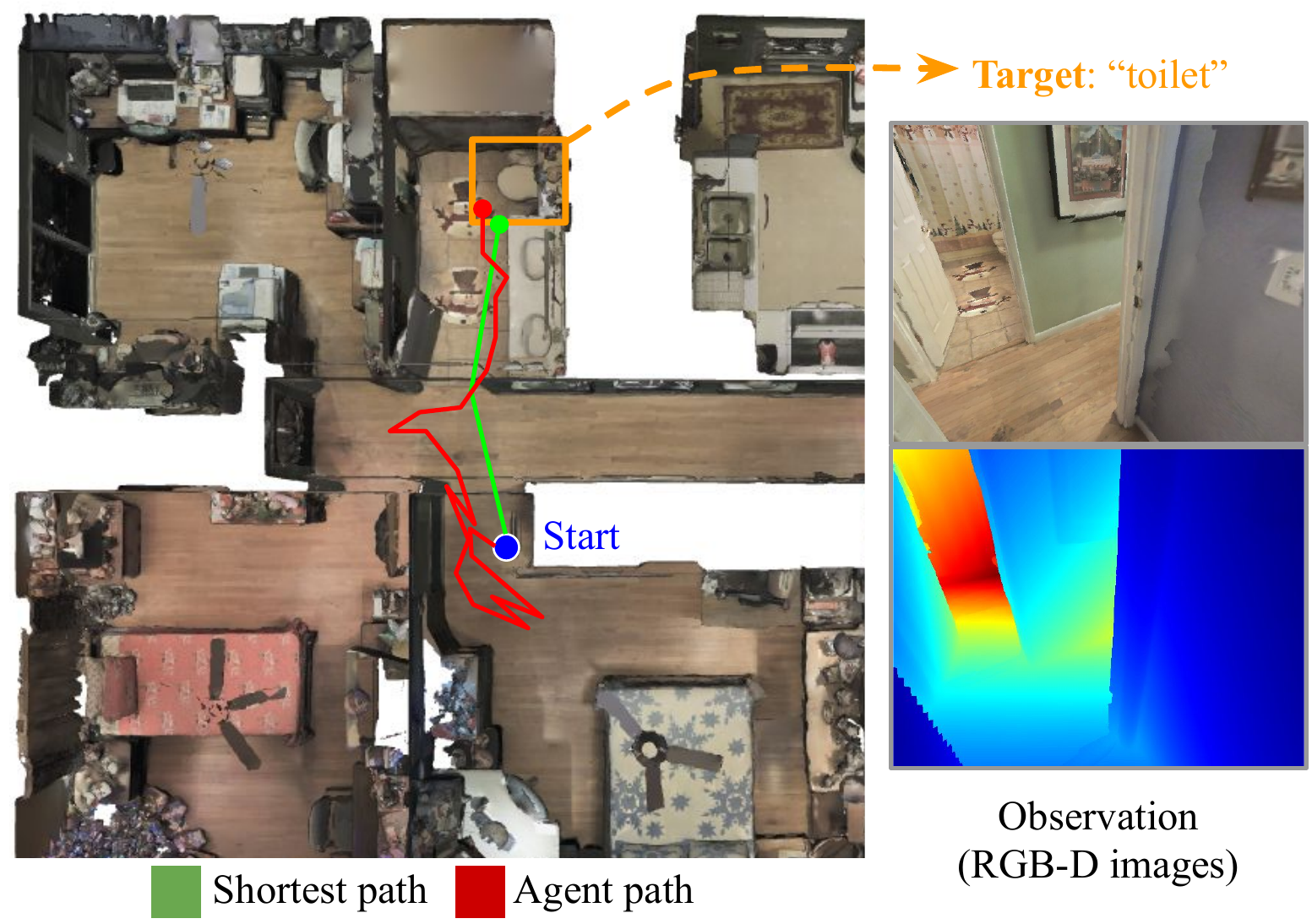} \vspace{-2mm}
\caption{\textbf{Visual Semantic Navigation.} The goal of the algorithm is to navigate an agent equipped with a first-person RGB-D camera as well as pose sensors from a random location in an \textit{unknown} scene to a given target object category (e.g., toilet). \vspace{-5mm}}
\label{fig:task}
\end{figure}

This paper tackles the task of \textit{visual semantic  navigation}, specifically ObjectNav \cite{batra2020objectnav}, where the goal is to navigate an agent (facing a random direction) from a random location in an unknown environment to a specified target object category (e.g., toilet) given first-person RGB-D image observations (shown in Fig. \ref{fig:task}).
In contrast to point goal navigation \cite{savva2017minos,xia2018gibson,yan2018chalet, Wijmans2019DDPPOLN}, where the goal location is provided as a local coordinate and the agent's only job is to find a collision-free path leading to that coordinate, this task requires the agent to simultaneously answer two questions: (1) where to go (i.e., the target object location) and (2) how to get there (i.e., planning an efficient path to reach the target object).


To efficiently find the target object with an unknown appearance in an unseen environment, the system needs to leverage its functional priors of typical indoor environments to guide its high-level search policy (e.g., beds are often located in bedrooms) as well as spatial prior for low-level action planning (e.g., how to exit this room without collision).
However, due to occlusion and limited camera field of view, the agent's observation of the environment contains only a small portion of the entire environment (e.g., a corner of a room in a big house). 
Thus this task requires a strong contextual prior of the complete 3D environment beyond an agent's partial observation.

In this paper, we introduce Confidence-Aware Semantic Scene Completion for Visual Semantic  Navigation (\textbf{SSCNav}), an algorithm that explicitly models this scene prior using a confidence-aware semantic scene completion module and uses this scene representation to guide the agent's navigation planning. 
Given a partial view of an indoor scene in the form of RGB-D images, the semantic scene completion module predicts the semantic labels for the complete environment (in a top-down map) centered around the agent.
By learning the statistics of many typical room layouts, the module is able to leverage contextual clues to predict what is beyond the field of view for typical indoor environments.  
Then, the prediction of the complete 3D environment is used by a learning-based navigation algorithm for generating next-step actions.  
Since the inferred scene is used for planning, it is important for the completion module to output confidence in its predictions.
To do so, the module produces a dense self-supervised confidence map together with the predicted environment map. Our experiment shows that the confidence estimation is critical for effectively guiding navigation and exploration by indicating to what degree the navigation policy could trust the inference result from the semantic scene completion module.  

The navigation module is trained with deep reinforcement learning \cite{qi2018human}, where the policy network takes in the agent's current state and target object category as input and outputs the next action. The state is represented by a completed semantic and confidence map, and the action is represented by a dense spatial action map \cite{wu2020spatial}. In this map, each pixel corresponds to an action that moves the agent towards that pixel's corresponding spatial location. In contrast to sparse action representation (e.g., steering command), this spatial action map representation is naturally aligned with our scene representation, and as we show in the ablation study, can significantly improve the performance \cite{wu2020spatial, zeng2018robotic,zeng2018learning,zeng2019learning}.


Our primary contribution is SSCNav, a framework that applies semantic scene completion with confidence estimation for the task of object goal visual semantic navigation. By leveraging the contextual priors of the typical indoor environment, our algorithm is able to infer a scene representation beyond its partial observation. We demonstrate that completed scene representation with confidence estimation enables more efficient planning policies for the downstream navigation task. Code and data will be available online.


\section{Related Work}

\mypara{Scene completion:}
SSCNav learns scene prior through scene completion, which is an idea explored in many recent works. 
Pathak et al. proposed ContextEnconder \cite{pathakCVPR16context}, which uses image completion to learn representations that capture contextual information. 
Song et al. proposed explicitly encoding 3D structure and semantic information in scene completion networks \cite{song2017semantic,song2018im2pano3d}.
However, these works do not study the effect of using these scene completion models in the context of motion planning.
Specifically, they do not provide a way to measure the uncertainty of the network prediction, which is critical for downstream motion planning algorithms. 
Jayaraman and Grauman \cite{jayaraman2016look,jayaraman2018learning} studied the use of panorama view completion for the task of selecting views. Ramakrishnan et al. \cite{ramakrishnan2019emergence}  demonstrated the effectiveness of such learned policies in a few action-perception applications. However, the task of image completion is performed in RGB color space without modeling semantic or 3D structure. 
Moreover, the actions used in these works are constrained to camera rotation and hence they cannot directly support navigation tasks that require translation actions.

\mypara{Point goal navigation:} 
Point goal navigation is a well-studied problem with many prior works \cite{savva2017minos,xia2018gibson,yan2018chalet, Wijmans2019DDPPOLN}.
In this task, taking in a target location and egocentric observations as input (images with color, depth, semantic segmentation, etc.), the agent needs to select one of the many possible actions and execute until it reaches the target coordinate. 
However, semantic goal navigation, specifically ObjectNav is much more challenging: since the target's location and appearance are both unknown to the agent, the system needs to estimate the target coordinate and the path towards it at the same time.

\mypara{Semantic goal navigation:} 
There have been rising interests in semantic navigation such as ObjectNav \cite{habitat19iccv}, where the goal is to navigate to a target object category instead of a global coordinate.
%
For example, the self-adaptive visual navigation method (SAVN) \cite{Wortsman_2019_CVPR} uses meta-learning to allow adaptation to unseen environments. Goal-Oriented Semantic Exploration (GOSE) \cite{chaplot2020object} builds a global top-down map and uses it to improve exploration efficiency. 
Value Learning from Videos (VLV) \cite{chang2020semantic} learns statistical regularities between object categories by watching YouTube videos. 
Another line of work utilizes semantic graphs to encode the scene structure for navigation \cite{yang2018visual, Qiu2020TargetDV}. The target goal is specified by text and the scene prior is learned with a language graph module which mostly focuses on object functional priors captured in language.
Different from these works, our algorithm explicitly captures the scene prior with a scene completion module, which naturally encodes the knowledge of both functional priors and object spatial relationships that may not be reflected in text or 2D videos.  
%

\begin{figure*}[!t]
    \centering
    \includegraphics[width=0.98\linewidth]{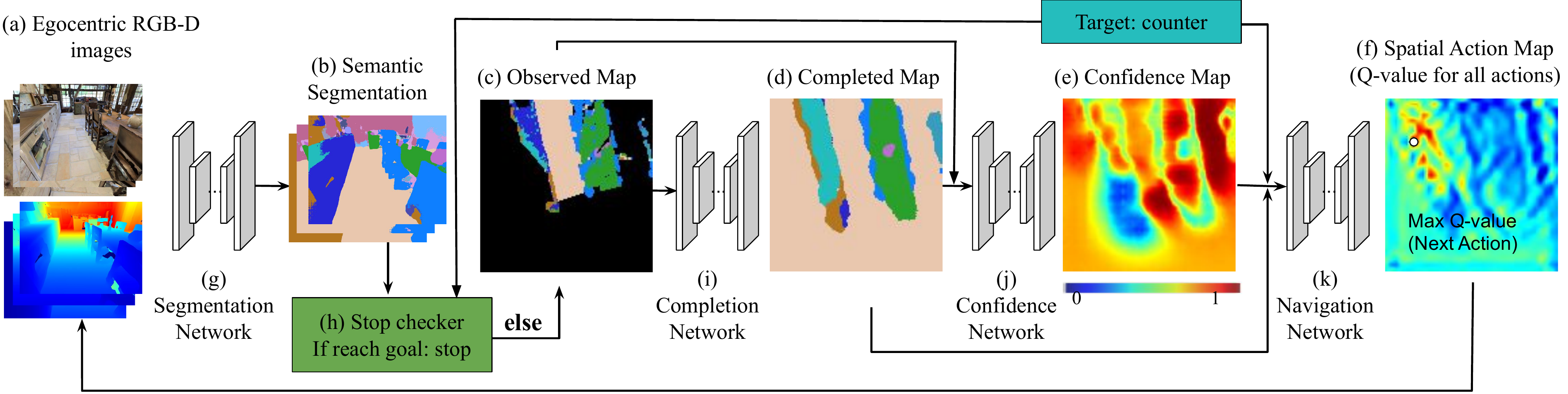} \vspace{-2mm}
    \caption{\textbf{SSCNav Overview.} At each step of an episode, the agent will get a new RGB-D image pair that is first used to predict the semantic segmentation (b) of the agent's current view. The past five observations are then aggregated and projected as a semantic top-down map (c), which is then completed by the scene completion network (i). The completed top-down map (d) and the observed semantic map (c) is then used by the confidence network (j) to estimate the confidence map (e) of the prediction. Next, the completed top-down map (d), the confidence map (e), and the target object category are used by the navigation network to predict a spatial action map (f), which is used to choose the next action. Finally, the stop checker (h) uses the new observation's semantic segmentation to determine whether the goal is reached. }
    \label{fig:overview}
    \vspace{-4mm}
\end{figure*}

\section{Approach}
Our approach consists of two major components: (1) a semantic scene completion module that takes in the agent's partial observation of the environment and infers a completed semantic representation $\hat{o}_t$ with confidence $c_t$, and (2) a navigation module that takes in the estimated $\hat{o}_t$, $c_t$, and the target object category to generate the next best action for the agent to move towards the goal. 
Fig. \ref{fig:overview} shows the overview of our approach. Sec. \ref{sec:completion} describes the confidence-aware semantic scene completion module and Sec. \ref{sec:navigation} presents the navigation module. 

\subsection{Confidence-Aware Semantic Scene Completion}
\label{sec:completion}

\mypara{Environment and agent setup:}
We use Habitat \cite{habitat19iccv} with Matterport3D environments \cite{Matterport3D} to train and test our algorithm. 
In our setting, the agent uses a first-person-view RGB-D camera with an image resolution of $480\times640$. The camera is at a height of $1.25\mathrm{m}$ from the floor, facing slightly downward at a $-30^{\circ}$ angle. At each step, the agent receives a new RGB-D image. For each image, we obtain a semantic segmentation using off-the-shelf ACNet \cite{hu2019acnet}. We train the ACNet with 209,200 RGB-D images of 40 object categories from Matterprot3D training houses.  

\begin{figure}[!t]
    \centering
    \includegraphics[width=0.98\linewidth]{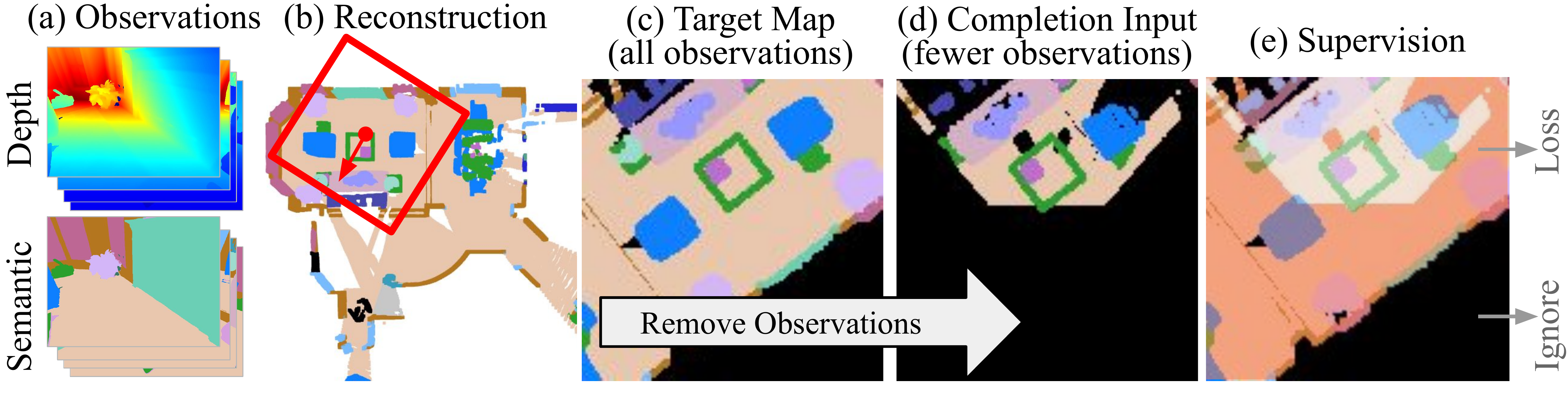} \vspace{-2mm}
    \caption{\textbf{Scene Completion Data.} The agent-centric depth and corresponding semantic observations are collected to obtain a global reconstruction of the 3D environment. Then the target semantic top-down map is generated with all the observations, and the input top-down map to the completion network is generated with fewer observation images (one to four views). To train the completion network, we only compute losses in regions that are observed in the target top-down map and ignore the unobserved regions. } 
    \label{fig:data}
    \vspace{-4mm}
\end{figure}

\mypara{Agent-centric semantic top-down map:} 
While the agent is moving in the environment, it aggregates the RGB-D observations, segmentation maps, and corresponding camera poses from the last $5$ steps. The camera poses are used to combine past observations into one 3D point cloud. Here the groundtruth camera poses are provided by the environment, following the setting of Habitat 2020 ObjectNav challenge \cite{batra2020objectnav}, as sensor pose estimation is out of scope for our task. After removing points that are higher than $1.55\mathrm{m}$ from the agent's standing floor or lower than the floor, the remaining point cloud is projected to an agent-centric semantic top-down map $o_t$, at step $t$. This top-down map represents a $6\mathrm{m}\times6\mathrm{m}$ local region centered around the agent and aligned with its orientation (Fig. \ref{fig:data} c). Each pixel on the top-down map is labeled with the corresponding semantic category, represented with a one-hot vector with $(N + 1)$ channels where $N$ is the number of object categories, and the extra channel indicates the unknown category.
This agent-centric semantic $o_t$ is input to the scene completion network.

\mypara{Scene completion network:}
The goal of the scene completion network is to take in the partial observation ${o}_t$ as input and output a completed top-down map $\hat{o}_t$ where the unobserved regions are filled with predicted semantic information. 
The completion module is implemented with a fully convolutional neural network with 1 maxpooling layer, 4 down-sample residual blocks, and 5 up-sample residual blocks \cite{He_2016_CVPR}. 
The network is trained to minimize the pixel-wise Cross-Entropy loss between the predicted semantic label and the groundtruth on the ``unobserved region''. 

\mypara{Self-calibrated confidence estimation:}
\label{sec:navigation}
The semantic scene completion module estimates the agent's surrounding environment based on typical distributions of indoor environments learned from the training data.
However, in practice, high variance in scene layouts and severe occlusion might lead to predictions with high uncertainty that may mislead the navigation module.

To mitigate this problem, We explicitly estimate the network's prediction uncertainty with another branch that outputs a corresponding confidence map $c_t\in [0,1]^{N \times M}$.
The confidence map branch has the same structure as the completion network, and is trained with self-supervision. By comparing the network prediction with the groundtruth, we obtain a target correctness map ${c_t}^g\in [0,1]^{N \times M} $, where each value denotes whether the prediction is correct or not. Then, the confidence network is tasked with predicting this pixel-wise correctness value based on the input observation $o_t$ and completed semantic map $\hat{o}_t$. The confidence network is trained to minimize the pixel-wise MSE loss between $c_t$ and ${c_t}^g$ on the ``unobserved region'' only. 

\mypara{Training:}
\label{sec:train}
To obtain training data for the scene completion network, we let the agent randomly explore an environment, aggregate its observations and estimations (e.g., color, depth), and reconstruct the 3D scene. During exploration, we train the scene completion network by randomly removing observed views. The network is tasked to infer the missing data. The unexplored region of the environment will be ignored for this training step (as shown in Fig. \ref{fig:data}). This learning scheme allows the system to learn from any new environment through navigation without fully reconstructed environments. 
In our experiment, we generated 52,300 groundtruth top-down maps for training, and each groundtruth map is paired with four different input observations with randomly removed views.

\subsection{Visual Semantic Navigation with Reinforcement Learning}\label{sec:navigation}

\mypara{DQN formulation:}
The navigation task is formulated as a Markov Decision Process. 
Given a target object category $g$ and current state $s_t$, the navigation policy $\pi(s_t)$ outputs the agent's next action $a_t$, then receives the next state $s_{t+1}$ and a scalar reward $r_t$ from the environment. 
The learning objective is to find a sequence of actions that maximize the total expected future rewards $ \Sigma_{i=t}^{\infty} \gamma^{i-t} r_i$, where $\gamma \in (0,1)$, a discounted sum over all future returns rewards.

We use off-policy Q-learning~\cite{mnih2015human} to learn a policy $\pi(s_t)$ that chooses actions which maximize a Q-function $\argmax_{a_t} Q_{\theta}(s_t,a_t)$ 
(i.e. state-action value function), which is a fully convolutional neural network \cite{He_2016_CVPR,long2015fully} parameterized by $\theta$.
We train our agents using the double DQN learning objective~\cite{van2016deep}. Formally, at each step $t$, the objective is to minimize: $\mathcal{L} = |r_t + \gamma Q_{\theta^-}(s_{t+1},\argmax_{a_{t+1}}{Q_{\theta}(s_{t+1},a_{t+1})})-Q_{\theta}(s_t,a_t)|$, where $(s_t,a_t,r_t,s_{t+1})$ is a transition uniformly sampled from the agent's replay buffer. Target network parameters $\theta^-$ are held fixed between individual updates and updated less frequently.


\mypara{State representation:} The state representation $s_t$ includes three parts: (1) the completed semantic top-down map $\hat{o}_t$, (2) the corresponding confidence estimation $c_t$, and (3) the target object category $g$. The target object category is encoded as a one-hot vector with $(N+1)$ channels ($N$ object categories plus background). This vector is tiled to have the same shape as $c_t$ and $\hat{o}_t$. 
We concatenate all three elements along the channel as the state representation.

\begin{figure*}[!t]
    \centering 
    \includegraphics[width=0.96\linewidth]{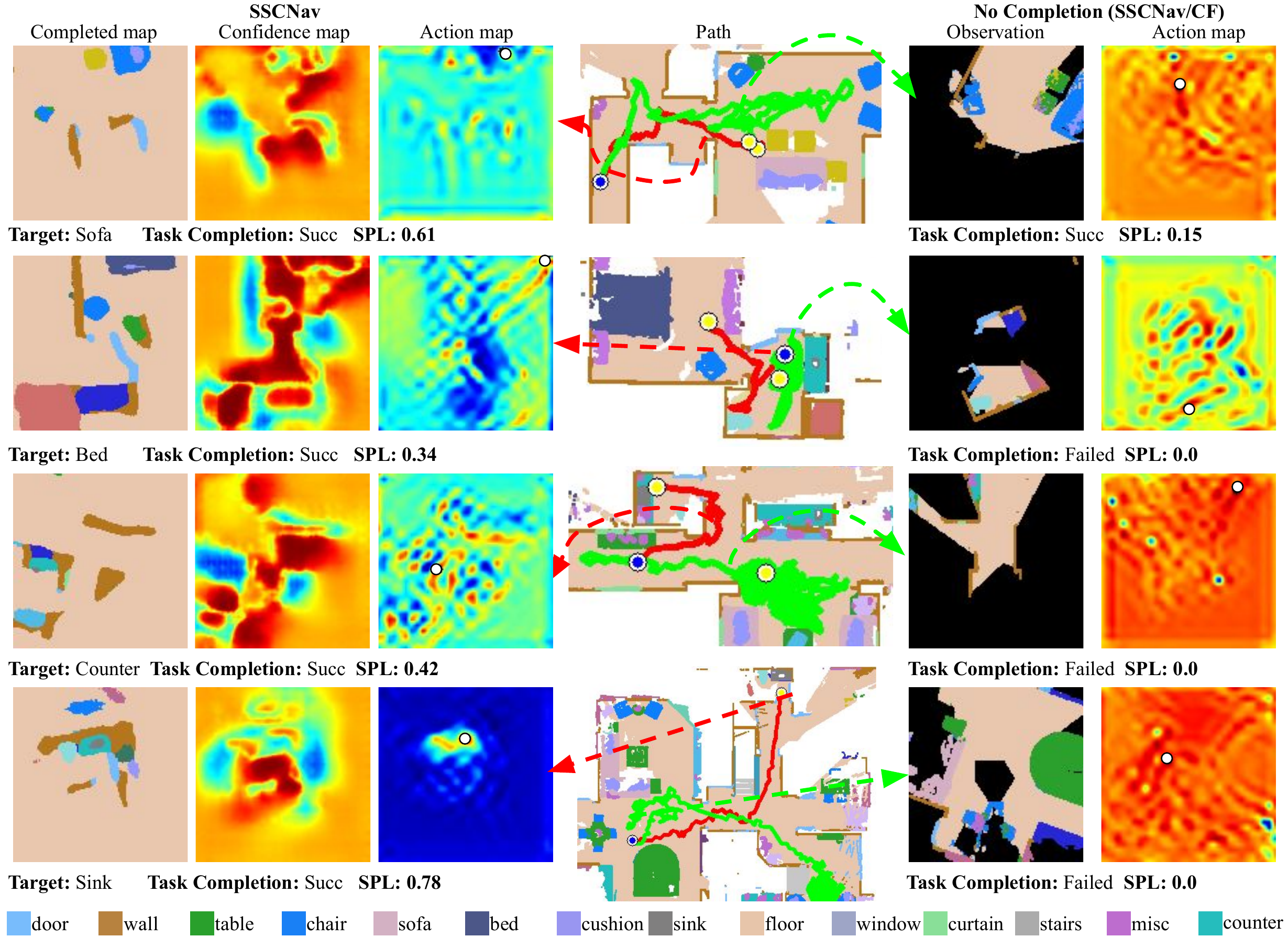} \vspace{-2mm}
    \caption{\textbf{Qualitative Result of Semantic Navigation.} In the path visualization (column 4), the red path shows the navigation path of the agent using the semantic scene completion module (SSCNav), and the green path is taken by the agent without semantic scene completion (SSCNav-CF). The blue dot indicates the start position and the yellow dot indicates the end position. Note that in the 1st row, agents from both approaches successfully reach their goals (a sofa) but SSCNav-CF makes a huge mistake and thus struggles to find the target. }
    \label{fig:nav_result_quali}
    \vspace{-5mm}
\end{figure*}

\mypara{Action representation:} 
We represent the action space $\mathcal{A}$ as a spatial action map similar to Wu et al. \cite{wu2020spatial}. The spatial action map representation shares the same spatial size as the state, where each pixel in the top-down map corresponds to a local navigational endpoint in the agent-centric map. At each step, after the agent selects a pixel location from the action map, it turns toward the corresponding point in the environment and moves forward for one step with a maximum step size of $0.25\mathrm{m}$ (collision is possible). 
To select the best action, the navigation network estimates the expected Q value for all the possible actions represented as a heat map (e.g., Fig. \ref{fig:nav_result_quali}). At test time, the policy picks actions by greedily choosing the pixel with the highest Q-value.  

This Q-value representation is spatially aligned and anchored to our scene representation, enabling significantly faster learning of complex behaviors in navigation, which is demonstrated in ablation study. 
Additionally, our action representation allows the Q-value of each local navigation endpoint and, thus, the behavior of the navigation policy, to be visualized on a heat map. 

\mypara{Reward:} At time step $t$, the agent gets a reward $r_t$ that is the sum of:

\begin{itemize}
 \item A life penalty of $-0.01$ to encourage a shorter path. 
 \item A penalty of $-0.25$ if the distance of the step is $< 0.125\mathrm{m}$, to encourage movement and punish collision. 
\item A delta-distance reward proportional to the change in the agent's distance to the closest target object.
\item A success reward of $+10$ if the agent reaches success.
\vspace{-2mm} 
\end{itemize}

\mypara{Stop checker:} As part of the task, the system needs to decide whether it has successfully reached its goal.  
To model this ability, we design a Stop Checker (Fig. \ref{fig:overview} h) that uses semantic and depth information to check whether the target object category is reached. 
From the agent's current standpoint, we rotate its camera along the z-axis 4 times and get pairs of observations. The Stop Checker returns true and terminates an episode if there are enough ($>5,000$) pixels belonging to the target object category within $1\mathrm{m}$ distance to the agent in at least 1 pair of observation. The episode will also be terminated if the agent has navigated for $500$ steps.

\mypara{Task completion:} At each action step, the environment will check whether the agent has succeeded and provide rewards accordingly. We use the following success metric according to \cite{batra2020objectnav}(note that validity is trivially met in our setting): (1) Intentionality: the agent's Stop Checker has returned true. (2) Proximity: the agent is within $1$ meter away from one target instance (3) Visibility: at least one target instance is visible. 

\mypara{Training:}
Each training/testing episode $\mathcal{T}$ contains 4 elements: a start location $p$, a start direction $f$, a scene $h$, and a target object category $g$.   
For a training episode $\mathcal{T}_k$, the target object categories are uniformly sampled. We then sample a training house from Matterport3D's training set and randomly choose a legal start point $p$. A start point is considered legal if it (1) is navigable (2) is in one of the valid room types (bathroom, bedroom, dining room, kitchen, living room, laundry room, and family room) to avoid starting outdoor, and (3) is far from all target objects (both the Euclidean and Geodesic distance to any target instance are greater than $r+\mathrm{D_{succ}}$, where $r$ is the instance's radius and $ \mathrm{D_{succ}}$ is success distance ($1\mathrm{m}$)).
If such a point is not found after $100$ trials, the search is repeated. Otherwise, the agent is turned to face a random direction and ready to start.

\section{Experiments}

\begin{figure}[!t]
     \centering
     
     \includegraphics[width=\linewidth]{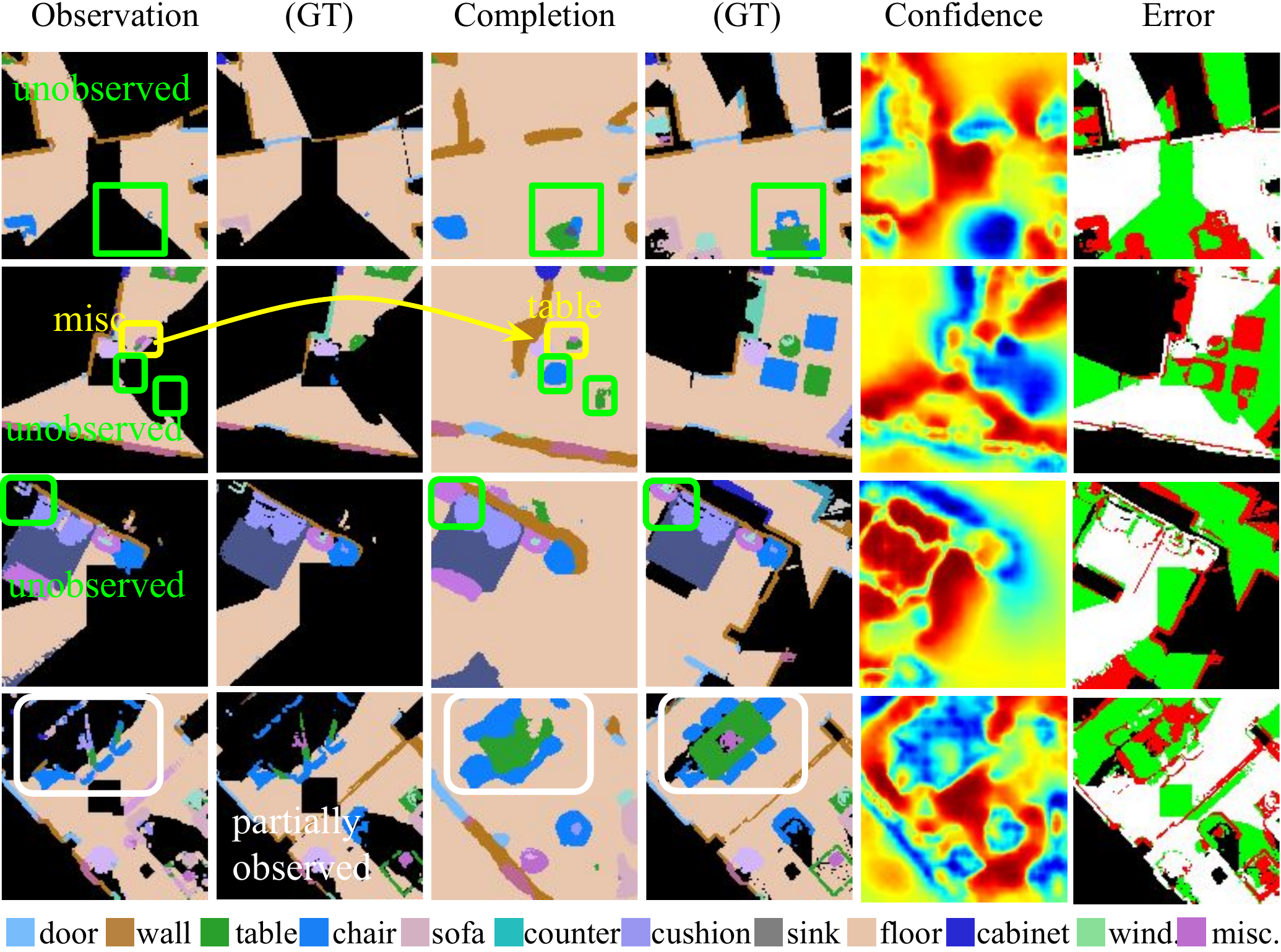}\vspace{-2mm}
     \caption{\textbf{Scene Completion Result.} From left to right are predicted and groundtruth segmentation of input observations, predicted and groundtruth scene completion, confidence estimation, and error map (white: observed area, red: incorrect completion, green: correct completion). The results show that the scene completion network is able to complete partially observed object (row 4), infer unobserved object (row 1,2,3), and correct segmentation error in the input observations (row 2). }
     \label{fig:cmplt_result_quali} 
\vspace{1mm}
    \centering
    \includegraphics[width=\linewidth]{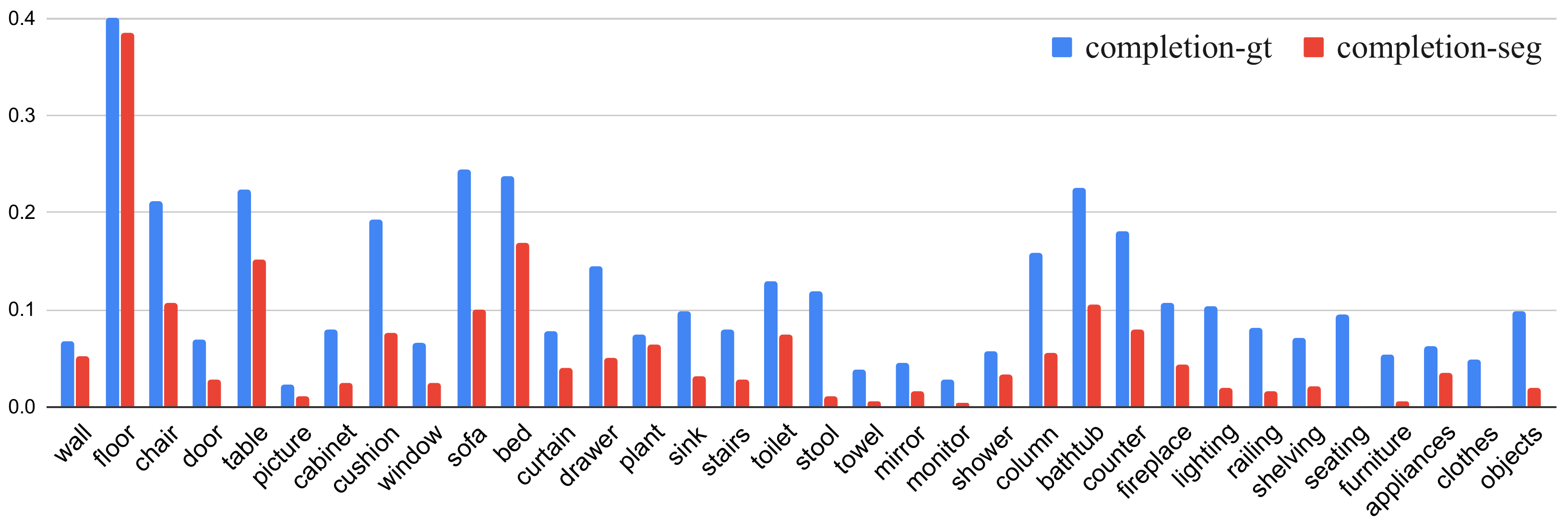} \vspace{-5mm}
    \caption{ \textbf{Semantic Scene Completion Results.} The plot shows intersection over union (IoU) of estimated semantic labels and groundtruth labels in the unobserved regions when taking groundtruth segmentation as input (blue) and estimated segmentation as input (red). Six categories (beam, blinds, gym-equip, board, ceiling, misc) that have zero IoU are not included.}
    \label{fig:cmplt_iou}
    \vspace{-5mm}
\end{figure}

\begin{table*}[t]
\caption{Navigation Result (Success Rate / SPL). /SA: using sparse action, -CF: without scene completion and confidence, -F: without confidence, /BC: binary confidence based on visibility. G+ using ground truth segmentation for observed scene. }
\setlength\tabcolsep{6 pt}
\centering
{
\begin{tabular}{l|ccccccc|c@{}}
\toprule
 Model            &     Bed       & Counter                & Shower & Sink       & Sofa          &  Table          & Toilet &  Avg                \\ \midrule
G+SSCNav-CF (GOSE \cite{chaplot2020object})  & 0.083/0.024 & 0.136/0.062 & 0.200/\textbf{0.101} & 0.207/0.098 & \textbf{0.377/0.182}  & 0.591/0.380 & 0.036/0.003 & 0.387/0.232\\
G+SSCNav-F  & 0.083/\textbf{0.061} & 0.121/0.054 & 0.043/0.015 & 0.268/\textbf{0.158} & 0.208/0.129 & 0.574/0.394 & 0.071/0.004 & 0.357/0.235\\
G+SSCNav  & \textbf{0.104}/0.057 & \textbf{0.227/0.092} & \textbf{0.257}/0.073 & \textbf{0.280}/0.149 & 0.245/0.115 & \textbf{0.656/0.425} & \textbf{0.143/0.061} & \textbf{0.438/0.259}\\ 
 \midrule
SAVN \cite{Wortsman_2019_CVPR} & 0.000/0.000 & 0.000/0.000 & 0.000/0.000 & 0.000/0.000 & 0.000/0.000 & 0.018/0.018 & 0.000/0.000 & 0.009/0.009 \\
SSCNav/SA & 0.000/0.000 & 0.015/0.002 & 0.000/0.000 & 0.049/0.019 & 0.000/0.000 & 0.206/0.086 & 0.000/0.000 & 0.109/0.045\\ 
SSCNav-CF (GOSE \cite{chaplot2020object})  & 0.021/0.002 & 0.015/0.005 & 0.043/0.009 & 0.037/0.027 & \textbf{0.151/0.055} & 0.394/0.219 & 0.000/0.000 & 0.218/0.118 \\
SSCNav-F & 0.042/0.035 & 0.076/0.034 & 0.000/0.000 & 0.122/0.028 & 0.019/0.009 & 0.385/0.216 & 0.000/0.000 & 0.217/0.117 \\ 
SSCNav/BC & 0.000/0.000 & 0.121/0.027 & 0.057/0.006 & 0.232/0.083 & 0.038/0.009 & \textbf{0.409/0.275} & 0.036/0.010 & 0.252/0.150 \\
SSCNav & \textbf{0.042/0.040} & \textbf{0.152/0.040} & \textbf{0.200/0.059} & \textbf{0.268/0.097} & 
0.057/0.029 & 0.388/0.261 & \textbf{0.107/0.037} & \textbf{0.271/0.157} \\ 
\bottomrule
    
\end{tabular}

\label{tab:nav_result}
}
\vspace{-4mm}
\end{table*}

In this section, we provide experimental results to validate our proposed approach. We run all experiments on the Habitat platform \cite{habitat19iccv} using the standard train-test split of Matterport3D \cite{Matterport3D}. In all the experiments, the agent is tested in \textit{novel houses}. Sec. \ref{sec:completion_result} summarize the result of semantic scene completion and Sec. \ref{sec:nav_result} study the effect of using scene completion in navigation. 

\subsection{Semantic Scene Completion}
\label{sec:completion_result}
\mypara{Metric:}
We evaluate the performance of scene completion with 715 samples from unseen houses in Matterport3D. We use the category-level pixel intersection over union (IoU) as the evaluation metric. The IoU is computed in the regions that are \textit{unobserved} region in the input top-down map but observed in the groundtruth \cite{song2018im2pano3d}. 

\mypara{Result:}
Fig. \ref{fig:cmplt_iou} summarizes the quantitative results of the scene completion network by using estimated semantic segmentation (completion-seg) and groundtruth semantic segmentation (completion-gt) as input observation. 
Fig. \ref{fig:cmplt_result_quali} shows the qualitative results and the estimated confidence map. 
The results show that the scene completion network is able to: 
(1) complete partially observed object geometry -- Fig. \ref{fig:cmplt_result_quali} row 4. 
(2) infer unobserved object based on contextual information -- Fig. \ref{fig:cmplt_result_quali} row 3 shows an example where the algorithm is able to infer the existence of an unobserved nightstand beside the bed based on the observed nightstand and infer the unobserved table and chair in row 1,2. 
(3) correct the segmentation error from the semantic segmentation network in the input observation -- Fig. \ref{fig:cmplt_result_quali} row 2 shows an example where the completion network corrects the incorrect prediction of misc to correct table class. 
Meanwhile, the completion model also produces incorrect prediction (labeled as red in Fig. \ref{fig:cmplt_result_quali}). The confidence score generated by the network, in general, reflects the confidence of the prediction (lower score for less-confident).   
In the next section, we will study the effect of using this semantic scene completion module in navigation.

\subsection{Semantic Navigation Result}

\label{sec:nav_result}
\mypara{Metric:}
We evaluate our models with 687 navigation episodes from unseen houses in Matterport3D provided by Habitat, with categories in \{Bed, Counter, Shower, Sink, Sofa, Table, Toilet\}.
We use the standard navigation success rate and SPL (Success weighted by normalized inverse Path Length) as metric \cite{Anderson2018OnEO}.
Note that there might be multiple object instances that belong to the target category. In this case, we will choose the closest target object when the episode starts to compute the shortest path. 

%

\mypara{Comparison with prior work:}
We evaluate and compare our approach to SAVN from Wortsman et al. \cite{Wortsman_2019_CVPR}. 
%
For the comparison, we finetune and test SAVN using the same setting as our method and report their performance in Tab. \ref{tab:nav_result} ([SAVN] v.s. [SSCNav]). Although SAVN only has live penalty and success reward during training originally, we add extra continuous distance reward (the same as our method's) to narrow the gap. The result shows that our method [SSCNav] significantly outperforms [SAVN] in all object categories.   
We conjecture that the reasons are two-fold. First, SAVN uses sparse action representation where $\mathcal{A}$ = $\{$ MoveAhead, RotateLeft,..., Done $\}$, and the mapping from observations to action classes is much harder to learn compared to the dense prediction enabled by the spatial action maps used in our method. Second, SAVN does not explicitly model semantic object classes in the scene representation, which might make the learning more difficult.  

%
Another prior work is GOSE \cite{chaplot2020object}, the winner of Habitat 2020 ObjectNav Challenge \cite{batra2020objectnav}. GOSE uses a similar top-down semantic map and spatial action representation. 
However, since the code is not publicly available, and the main difference is the lack of semantic scene completion module, one could refer to the [SSCNav-CF] performance (SSCNav without (-) scene (C)ompletion  and con(F)idence ) in Tab. \ref{tab:nav_result}, as our attempt of re-implementing their system.

\mypara{Does confidence-aware scene completion help?}
To test the effect of using semantic scene completion module in navigation, we compare the navigation performance of models trained with and without the semantic scene completion module ([SSCNav-CF] v.s. [SSCNav-F] v.s. [SSCNav], C, scene (C)ompletion F, con(F)idence). 
By comparing [SSCNav-CF] and [SSCNav-F], we observe that directly using the scene completion does not affect the navigation success rate. We conjecture that the error and noises in the scene completion result can still be misleading for navigation planning, and the confidence information naturally carried in the scene completion output (after softmax) is not sufficient.

This problem is largely mitigated with the confidence estimation. [SSCNav] outperforms [SSCNav-CF] with +5.3\% in success rate and +3.9\% in SPL. Even when completion errors exist, [SSCNav] could recovers and outperforms [SSCNav-CF] in most categories in light of confidence awareness. 
For some categories that [SSCNav] is not outperforming [SSCNav-CF] the SPL numbers are much higher (e.g., table -0.6\% success rate, +4.2\% SPL). 
The result indicates that the confidence-aware semantic scene completion module is particularly helpful for objects with strong contextual bias (e.g., toilets only appear in restrooms). 
On the other hand, for common objects such as sofas and tables that can appear in almost any room, the contextual bias is less useful.

\mypara{What does confidence estimation learn?}
Given that confidence estimation is critical to allow the model to benefit from scene completion, we wonder whether our confidence estimation is better than naively masking seen area as `confident' and unseen area as `unconfident'? To answer this question, we include an additional baseline [SSCNav/BC] (BC, binary confidence), 
where the confidence is modeled by scene visibility: seen region with value 1, unseen region with value 0.   
The result shows that [SSCNav] outperforms [SSCNav/BC] +1.9\% in success rate and +0.7\% in SPL. 
This result validates that while the scene visibility is highly correlated with network uncertainty, the confidence estimation still carries more useful information. 



\mypara{Effect of spatial action representation.}
To test the benefit of using spatial action map as our action representation, we compare [SSCNav] with [SSCNav/SA] (/SA, spares action). [SSCNav/SA] is the replaces navigation model's output map with a sparse action probability set, a common strategy for such policy. We use the following action space: \{Turn right for $k$ degrees, then move forward  $ | k \in [0, 45, 90, 135, 180, 225, 270, 315]$\}. At each time step, the agent picks the action with maximum probability instead of the pixel with maximum Q-value. Along with losing the interpretability of the navigation model, we witness a significant drop both in success rate (-16.2\%) and SPL (-14.8\%). This result is consistent with prior work \cite{wu2020spatial}.

\mypara{Performance oracle.}
Here we evaluate the navigation performance when the groundtruth segmentation for the observed region is available.  
Tab. \ref{tab:nav_result}, [G+SSCNav]s show the navigation performance. 
While groundtruth input clearly improves the algorithm performance, the result still follows a similar tendency: even with perfect semantic segmentation as input, semantic scene completion still improves navigation performance.
This result indicates that with the advancement of general semantic segmentation algorithms, our system's performance will also be improved.

\section{Conclusion}

We introduce SSCNav for the task of visual semantic navigation. The algorithm leverages a confidence-aware semantic scene completion module to achieve a better understanding of the agent's surrounding environment and uses this scene completion result to facilitate its action planning in navigation. 
Our experiments demonstrate that by learning the contextual priors of the typical indoor environment, SSCNav is able to improve the performance of visual semantic navigation in terms of both success rate and efficiency. 


\bibliographystyle{IEEEtran}
\bibliography{IEEEabrv, mybibfile}

\end{document}